\documentclass[runningheads]{llncs}

\usepackage{eccvabbrv}
\usepackage{algorithmic}
\usepackage{algorithm}

\usepackage{amsmath}

\usepackage{caption}
\usepackage{graphicx}
\usepackage{booktabs}

\usepackage[accsupp]{axessibility}  

\usepackage[pagebackref,breaklinks,colorlinks,citecolor=blue]{hyperref}

\usepackage{orcidlink}

\begin{document}

\title{Attack $GAN$ ($AGAN$): A new Security Evaluation Tool for Perceptual Encryption} 


\author{Umesh Kashyap, Sudev Kumar Padhi and Sk. Subidh Ali}
\institute{Indian Institute of Technology Bhilai}

\maketitle

\begin{abstract}
Training state-of-the-art ($SOTA$) deep learning models requires a large amount of data. The visual information present in the training data can be misused, which creates a huge privacy concern. One of the prominent solutions for this issue is perceptual encryption, which converts images into an unrecognizable format to protect the sensitive visual information in the training data. This comes at the cost of a significant reduction in the accuracy of the models. Adversarial Visual Information Hiding ($AVIH$)  
overcomes this drawback to protect image privacy by attempting to create encrypted images that are unrecognizable to the human eye while keeping relevant features for the target model. In this paper, we introduce the Attack $GAN$ ($AGAN$) method, a new Generative Adversarial Network ($GAN$)-based attack that exposes multiple vulnerabilities in the $AVIH$ method.  To show the adaptability, the $AGAN$ is extended to traditional perceptual encryption methods of Learnable encryption  ($LE$) and Encryption-then-Compression ($EtC$). Extensive experiments were conducted on diverse image datasets and target models to validate the efficacy of our $AGAN$ method. The results show that $AGAN$ can successfully break perceptual encryption methods by reconstructing original images from their $AVIH$ encrypted images. $AGAN$ can be used as a benchmark tool to evaluate the robustness of encryption methods for privacy protection such as $AVIH$.

\end{abstract}

\section{Introduction}
\label{sec:intro}
The recent development in deep learning has led to its wide applications. From image recognition to autonomous driving, almost every walk in our daily life is benefitted by the introduction of deep learning~\cite{he2016deep,simonyan2014very,baevski2020wav2vec,vaswani2017attention}. Usually, training deep learning models requires significant computational resources, making it infeasible to train on local machines, such as laptops and desktops. This is where cloud services come into the picture, offering scalable and efficient platforms to deploy, manage, and scale deep learning models and their applications. Deep learning models are required to be trained on a large amount of data sets. Oftentimes, these data sets consist of private data, such as medical images, facial and identity informations, $etc.$ Sharing privacy-critical data with the cloud service provider may lead to the risk of privacy violation~\cite{arockiam2014efficient,singh2017cloud,chen2012data}.  
Therefore, deploying solutions for the privacy protection of cloud data is an urgent need.

One of the prominent solutions for image privacy protection is visual information encryption. This method converts images into an unrecognizable format to keep the image content hidden. Visual encryption often requires protecting visual information such that the data is effectively encrypted without affecting its intended application. There are two popular methods in visual information encryption:  homomorphic encryption ($HE$) and perceptual encryption ($PE$). In $HE$~\cite{he4,he5,he6,he7},  deep learning models (service models in the cloud) are converted to work on encrypted data to perform tasks such as object detection, classification, $etc.$ Whereas in $PE$~\cite{pe1,pe2,pe3,pe4,pe5,pe6,pe7,pe8,tanaka,inverse}, spatial features of the image are used for the encryption. 
$HE$ methods are computation intensive and also not suitable for large non-linear computations such as deep learning models~\cite{pe6}. Therefore, $PE$ is the most suitable choice for deep learning-based visual encryption. 

$PE$ methods use features of an input image to transform it to the encrypted domain~\cite{pe2,pe3,pe4,pe5,pe7,pe8,tanaka} such that the service model can be trained on data in the encrypted domain. Nevertheless, the model accuracy drops substantially on the encrypted domain in comparison to the original domain.  To overcome the issues, a more advanced information transformation network for $PE$ is proposed~\cite{pe1,inverse} with the aim of increasing the service model's accuracy in the encrypted domain.  However, once encrypted, this method can not recover the original image~\cite{pe1,inverse}, thus limiting its functionality. In this line, the state-of-the-art ($SOTA$) deep learning-based $PE$ method for protecting the privacy of visual data is considered to be $AVIH$~\cite{pe6}. This method not only provides high validation accuracy of the underlying service model on the encrypted domain but also facilitates the recovery of images from the encrypted domain. $AVIH$ leverages $type 1$ adversarial attack to generate an encrypted image (perturbation) from the original image, which is feature-wise closer to the original image, while preserving a negligible reconstruction loss corresponding to a given conditional $GAN$. The security/privacy of this method lies in the secrecy of the initial seed of the $GAN$.   
 
The security of any encryption method lies in its robustness against existing attacks. In this line, the vulnerabilities of the traditional $PE$ methods introduced in~\cite{tanaka} and~\cite{pe5} are shown in~\cite{cipat}, where the adversary could easily unravel the relation between encrypted and unencrypted images, $i.e.,$ the shuffling and reversing the operation of the image encryption. Advanced information transformation network-based $PE$ is broken by utilizing an inverse-transformation network attack. This attack uses the original and encrypted images to train inverse-transformation network($U$-$Net$) for reconstructing the original image for a given encrypted image~\cite{pe1,inverse}. Later on, it is shown in~\cite{GAN_A1,GAN_A2} that a $GAN$ can be trained to unscramble the scramble-based encryption methods~\cite{pe8,pe4}. These attacks clearly show vulnerabilities in existing $PE$ methods, where the adversary may not recover the entire content from the encrypted image instead, it can recover certain attributes of the contents, resulting in a violation of privacy. In this line, the $SOTA$ deep learning-based $PE$ method $AVIH$~\cite{pe6} is designed to withstand these exiting attacks side-by-side, allowing authorized users to recover the original images from the encrypted images.

In this paper, we perform a detailed security analysis of $SOTA$ deep learning-based $PE$ method, $AVIH$. We propose a new security evaluation method referred to as $Attack$ $GAN$ ($AGAN$) for deep learning based $PE$ method. Our $AGAN$  shows two key vulnerabilities in $AVIH$: 1) Variance consistency loss does not sufficiently reduce the correlation between original and encrypted images. 2) Keeping the $GAN$ a secret is not secret enough. An adversary can train a surrogate $GAN$ to unravel its secrecy. 
Our key contributions are as follows:
\vspace{-0.2cm}
\begin{itemize}
    \item  We propose $AGAN$, a new security evaluation method for deep learning-based $PE$ which identifies multiple vulnerabilities in the $SOTA$ $PE$ method $AVIH$. 
    \item We conduct extensive experiments to show that $AGAN$ can easily recover the original image from the encrypted image that is encrypted with $AVIH$.
    \item We also demonstrate that $AGAN$ is task-specific service model independent.  In other words, $AGAN$ designed to attack one service model can be easily transferable to other service models with similar tasks.
    \item We demonstrate that $AGAN$ outperforms existing $SOTA$ deep learning-based attack methods to reconstruct the original image from the corresponding encrypted image.
\end{itemize}

\section{Related Works}
\subsection{Perceptual Encryption}
$PE$ protects the visual information of an image by transforming it into an encrypted form such that the required task can be performed on the encrypted image while keeping its visual content private.  $PE$ is the most prominent visual encryption method that works by preserving image intrinsic properties in the encrypted image~\cite{tanaka}. 
9Tradition $PE$ methods use block-wise and pixel-wise scrambling methods to create the encrypted image from the original image. Subsequently, the service model is trained on encrypted images to perform required tasks such as classification, object detection, $etc$. $LE$  method proposed in~\cite{tanaka} uses a deterministic encryption method that operates on color components within the blocks of an image. It involves shuffling and reversing the color components using a secret encryption key. In the same line, the encryption method in~\cite{pe5} is based on a probabilistic encryption method that uses a different encryption key for each image. It involves negative-positive transformations and optional shuffling of color components based on pseudorandom bits generated from the key. Similarly, $EtC$~\cite{pe8} method applies block rotation, inversion, negative-positive transform, color shuffling, and block shuffling with the same key per block. It is quite natural that training a deep neural network on encrypted images will be difficult in comparison with training on unencrypted images~\cite {GAN_A2}. To overcome this issue, a deep learning-based $PE$ method is proposed in~\cite{pe1,inverse}, which uses a transformation network that is trained to transform plain images into visually protected encrypted images while preserving the features required by the service model for successfully performing it's task. The key advantage of this method is that, unlike traditional methods, the service model does not need to be trained on encrypted images.  The only limitation of this method is that images can not be recovered once encrypted. Therefore, in this line, $AVIH$~\cite{pe6} is a $SOTA$ deep learning-based $PE$ method that successfully transforms the original image into an encrypted image while preserving sufficient features of the visual content of the image such that the service model can perform its tasks without retraining. The security of this method is based on two key features: variance consistency loss and using a $GAN$ as the secret key. Variance consistency loss helps in reducing the correlation between the original and encrypted images. Meanwhile, the secret $GAN$ reduces the chance of reconstruction from the encrypted image. Without this $GAN$, the adversary will not have access to the original image and thus can not train another $GAN$ for reconstruction.  Only authorized users possessing the secret $GAN$  can recover the original image from the encrypted image. Although $AVIH$ can overcome multiple lacunas of existing $PE$ methods, there is no independent security validation of $AVIH$. Therefore, the question remains: can a surrogate $GAN$ be trained to replicate the behavior of the secret $GAN$ of  $AVIH$?

\subsection{Vulnerabilities in Perceptual Encryption}
There are existing chosen-plaintext and ciphertext-only attacks proposed to break $PE$ methods. These attacks primarily focus on traditional $PE$ methods.  The initial attack~\cite{cipat} targeted the traditional methods of $LE$~\cite{tanaka} and~\cite{pe5}. The attack is based on creating a set of helper images $i.e$, creating a set of original images and their corresponding encrypted image. These helper images are of a specific size so they can determine precisely what shuffling and reversing has been done. This information is then used to decrypt the encrypted images. However, this attack will not work when almost every pixel in the image has the same correlation among each image block~\cite{GAN_A2}. 
An advanced attack called the inverse transformation network attack has been proposed against transformation network-based (PE) methods~\cite{pe1,inverse}. The inverse transformation network trains a U-Net architecture to reconstruct original images from their encrypted image. It learns the inverse mapping from encrypted image to original images using paired image datasets. The inverse network aims to invert the transformation network's visual encryption by modelling the inverse transformation. 
However, the limitation of the inverse transformation network attack is that it can only be effective when the encrypted images still retain some visual information or statistical correlation with the original image. Suppose the transformation network successfully removes all visual content corresponding to the original images during encryption, in that case, the inverse network will fail to reconstruct the original images from the encrypted images accurately.
$SIA$-$GAN$ solves this issue by proposing a method to decrypt the image, encrypted by the scramble-based method using  $GAN$ for the blockwise reconstruction of the original image~\cite{GAN_A2}. The efficacy of this attack is shown by successfully exploiting the encryption method of $EtC$~\cite{pe8,pe4}. Recently proposed $GAN$-based attack~\cite{GAN_A1} reconstructs original images from encrypted images generated through $Etc$~\cite{pe8,pe4}, with the aim of retrieving certain attributes of the visual content. However, neither of these $GAN$-based attacks can completely reconstruct the original content of encrypted images. To overcome this issue, Our $AGAN$ method takes a step further and successfully extracts the original image from a given encrypted image. The generated original image contains almost all the original content's attributes, making it easy to identify the content. Unlike previous methods, $AGAN$ does not require any information about the original image $x$ while generating it from the corresponding encrypted image $x^\prime$. $AGAN$ successfully attacks traditional and deep learning-based $PE$ methods. To show its effectiveness, we have performed an attack on $ETC$~\cite{pe8}, $LE$~\cite{tanaka}, and the $SOTA$ deep learning based $PE$ method $AVIH$.

\section{Proposed Method}
$AGAN$ method is developed targeting $SOTA$ deep learning-based $PE$ method $AVIH$. Later, we will show that this method can be extended to other $PE$ methods with little modifications. 

\subsection{Threat model}
The target $PE$ method ($AVIH$) is based on a trained service model $f$ and a generator $G$ (secret model). The encrypted image $x^\prime$ is generated from the original image $x$ based on the following optimization problem:

\begin{equation}
\underset{x^\prime}{\text{minimize}} \: \mathcal{L}_f(f(x^\prime), f(x)) + \mathcal{L}_g(G(x^\prime), x) + \mathcal{L}_v(x^\prime)
\end{equation}
\noindent where, $L_f$ and $L_g$ are the feature and reconstruction loses, respectively, between the original image $x$ and the encrypted image $x^\prime$ with respect to the service model $f$ and secret model $G$ respectively. $L_v$ is the variance consistency loss calculated between the different blocks of $x^\prime$. The security of this method lies in the secrecy of the initial seed value of $G$. For more details on the $AVIH$ $PE$ method, one can refer to~\cite{pe6}. 

The adversary aims to recover the original image $x$ from the visually encrypted image $x^\prime$. The following are the assumptions about the adversary's knowledge:
The adversary 
\begin{enumerate}
 \item Has access to $f$ and $x^\prime$.  $f$ and $x^\prime$ both are in the cloud, where the cloud service provider is untrusted or the adversary itself. This is the key assumption, $i.e.,$ cloud services are untrusted, based on which all the $HE$ and $PE$ methods are developed
 \item knows the architecture of  $G$. It is also the assumption of the target $PE$ method 
 \item knows the $PE$ algorithm $E$
 \item does not have access to $G$ and its initial seed
 \item does not have access to original images $x$ corresponding to encrypted images $x^\prime$ stored in the cloud
\end{enumerate}

\subsection{Key challenges in attack }
 There are attacks on the $PE$-based method using $GAN$~\cite{GAN_A1,GAN_A2}. Here, the attacks work if the adversary has knowledge of the input data distribution used to train the original model. The adversary uses it to create its own data set pair of original and encrypted images, using a $GAN$-based model ($StyleGAN$ and  Diffusion model), which is trained to map the encrypted image to the original image.  In the case of $AVIH$, even if the adversary knows the data distribution along with the encryption algorithm, and the architecture of  $G$ and $f$, it will not be able to recover the original image because the encrypted image is highly dependent on the initial seed of $G$. Training a surrogate $GAN$ with a different seed will lead to a different local optima~\cite{pe6} than $G$.  Along the same line, adversarial perturbation is generated for performing $PE$ and to make the perturbation indistinguishable, variance consistency loss is used such that each part of the encrypted image has the same variance. This increases the dependency on the initial seed and makes it harder for the surrogate models to extract visual information. We proposed a novel $AGAN$ method to overcome this challenge. In $AGAN$, the master key model $G_a$ is trained into two sub-branches, where the first branch acts as a simple generator to learn reconstruction from  $\hat{x}^\prime$ and another branch overcomes the feature loss between reconstructed image $\widetilde{x}$ and original image  $\hat{x}$. The surrogate data was utilized to train the $G_a$ model, where pairs of original  $\hat{x}$ and encrypted  $\hat{x}^\prime$ images were collected from the encryption method. The detailed discussion for the surrogate data collection process is explained in Section~\ref{experiment}. $AGAN$ utilize both branches to learn the relation between pixels and reconstruct $\hat{x}$ for $\hat{x}^\prime$.

\begin{figure}[!htb]
  \centering
  \includegraphics{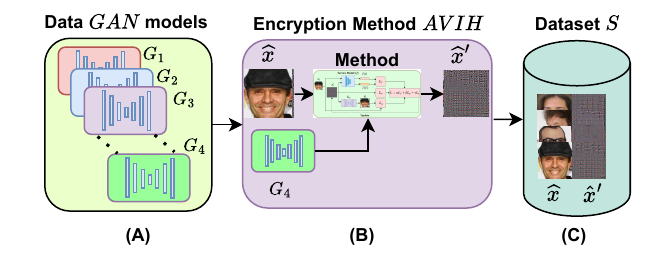}
  \caption{Overview of collecting dataset $S$ to train master key model $G_a$. (A) data $GAN$s model train with different seeds to learn reconstruction for original $\hat{G}(\hat{x})\rightarrow \hat{x}$. (B) Encryption method to take each of trained data $GAN$ model as a secret $GAN$ model and generate $N$ number of sample pair of ($\hat{x},\hat{x}^\prime$) (C) Collected different seed-based data pair samples in Dataset $S$ to train master key model $G_a$.
  \label{fig:2}
  }
\end{figure}

\subsection{Overview}

As a first step of the attack, we train four data $GAN$ models ($\widetilde{G_1}$,$\widetilde{G_2}$,$\widetilde{G_3}$,$\widetilde{G_4}$), where $\widetilde{G_1}$ and $\widetilde{G_2}$ are trained with $CelebA$ dataset while $\widetilde{G_3}$ and $\widetilde{G_4}$ are trained with $COCO$ dataset. It is to be noted that all four data $GAN$s are initialized with different seeds. We follow the same architecture for these four $GAN$ models as used for $G$ of $AVIH$. 
Once trained, each $GAN$s generate $10K$ pair of original image $\hat{x}$ and encrypted image $\hat{x}^\prime$. We then collect these to create a surrogate dataset $S$, which is in the form of ($\hat{x}$, $\hat{x}^\prime$). This $S$ is used to train our master key model $G_a$ as shown in Figure~\ref{fig:1}. The master key model consists of a generator $G_a$, discriminator $D$ and pre-trained model $h$, which acts as the feature extractor. In the $AGAN$ method, the generator generates image $\widetilde{x}$ for a given  $\hat{x}^\prime$ and the discriminator is trained simultaneously with the generator to discriminate between the $\hat{x}$ and the generated image $\widetilde{x}$. The working of both $D$ and $G_a$ are explained below.

\subsubsection{Discriminator:}  The Discriminator is tasked with distinguishing between original image $\hat{x}$ and generated image $\widetilde{x}$, derived from encrypted image $\hat{x}^\prime$. This model utilizes a $PatchGAN$ discriminator, which differs from the standard $ CNN$-based discriminator. Instead of considering the full image at once, it focuses on smaller parts known as patches. This allows the model to notice minute details and accurately estimate the image quality. Using standard discriminators can overlook minute features, particularly those that make images appear real. However, $PatchGAN$ focuses on these characteristics in different image patches, ensuring they are preserved for better image reconstruction. Its primary objective is to maximize its discriminatory accuracy, to accurately classify  $\hat{x}$ as real and reconstructed images $\widetilde{x}$ as fake as shown in Eq~\eqref{eq1}. 
\begin{equation}
\mathcal{L}_{adv} = \mathbb{E}_{\hat{x}} [\log D(\hat{x})] + \mathbb{E}_{\hat{x}^\prime} [\log (1 - D(G_a(\hat{x}^\prime)))]
\label{eq1}
\end{equation} 
$D$ tries to maximise the loss, while $G_a$ attempts to minimise the loss. This causes $D$ to accurately discriminate the original as real and the reconstructed as fake.

\subsubsection{Generator:} In our case, the generator needs to generate the original image $\hat{x}$ from visually encrypted image $\hat{x}^\prime$. The generator architecture of the master key model comprises two distinctive branches, each serving distinct purposes. The first branch is the generator $G_a$ itself, which is rigorously trained to generate the reconstructed image $\widetilde{x}$ corresponding to the visually encrypted image $\hat{x}^\prime$. Concurrently, the second branch incorporates a pre-trained model $h$ to extract features from the  $\hat{x}$ and $\widetilde{x}$. This strategy of branching the generator aims to enhance the generator's reconstruction capabilities. Overall, the loss function of our generator $G_a$ consists of two parts: $L_g$ for generating loss between ($\hat{x}$ and $\widetilde{x}$) as shown in Eq~\eqref{eq2} and $L_h$ which is the mean square loss between ($h(\hat{x})$,$h(\widetilde{x})$) to minimize the feature distance between  $\hat{x}$ and  $\widetilde{x}$ as shown in Eq~\eqref{eq3}. These are expressed following as:

\begin{figure}[!htb]
  \centering
  \includegraphics[scale=0.55]{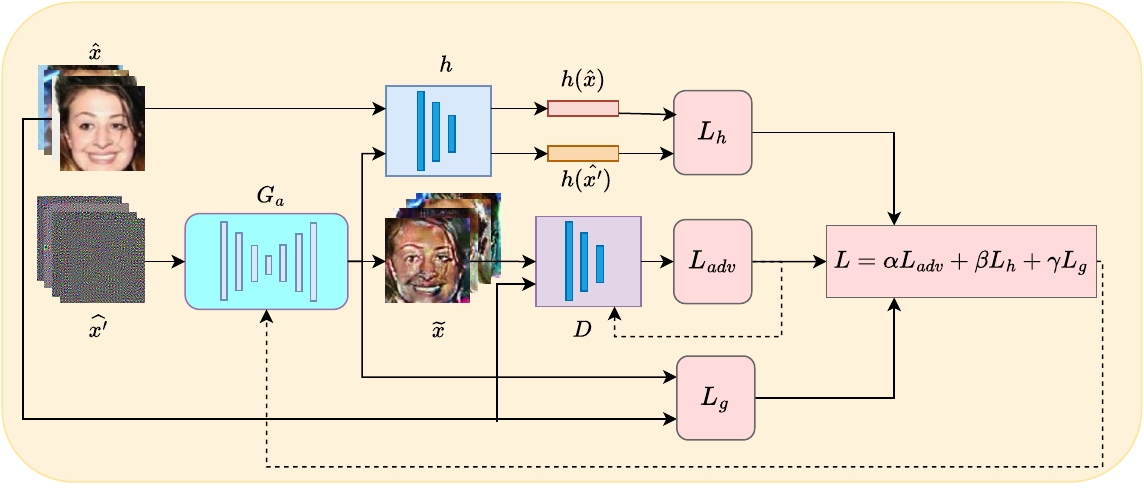}
  \caption{Overview of the proposed method to train the master key model $G_a$}. 
  \label{fig:1}
\end{figure} 

\begin{equation}
\mathcal{L}_{g} = \mathbb{E}_{\hat{x}^\prime,\hat{x}} [\|\hat{x} - G_a(\hat{x}^\prime)\|_1]
\label{eq2}
\end{equation} 

The generator network $G_a$ is trained to minimize $L_g$ loss, ensuring the reconstructed images $\widetilde{x}$ closely resemble the original images $\hat{x}$. Concurrently, $G_a$ aims to minimize its adversarial loss $\mathcal{L}_{adv}$ against the discriminator network $D$, effectively learning to generate realistic replicas of the  $\hat{x}$ from their $\hat{x}^\prime$, encrypted with $AVIH$.
\begin{equation}
    \mathcal{L}_{h} = \frac{1}{n} \sum_{i=1}^{n} [h(\hat{x}) - h(G_a(\hat{x}^\prime))^2]
    \label{eq3}
\end{equation}

\begin{algorithm}[!htb]
\caption{Training master key model $G_a$}
\label{algo:agan_model}
\begin{algorithmic}[1]
\REQUIRE
 Surogate dataset $S$, pre-trained feature extractor model $h$, number of iterations $K$, discriminator loss coefficient $\alpha$, generator loss coefficient $\beta$, feature loss coefficient $\gamma$.
\ENSURE Trained master key model $G_a$
\STATE Initialize generator $G_a$, discriminator $D$, and pre-trained feature extractor model $h$
\FOR {$k = 0$ to $K$}
    \STATE  $L_{\text{adv}} = E_{\hat{x}^[\log D(\hat{x}^)]} + E_{\hat{x}^\prime}[\log(1 - D(G_a(\hat{x}^\prime)))] $, where ($\hat{x}, \hat{x}^\prime \in S$)
   \STATE $L_g = E_{\hat{x}^\prime},\hat{x}^[||\hat{x}^ - G_a(\hat{x}^\prime)||_1] $
    \STATE $L_h = \frac{1}{n} \sum_{i,j=1}^{n} [h(\hat{x}) - h(G_a(\hat{x}^\prime))]^2 $
    \STATE $L = \alpha \cdot L_{\text{adv}} + \beta \cdot L_g + \gamma \cdot L_h $
    \STATE $G_a = G_a - \eta \cdot \nabla G_a(L)$ //Minimize $L$ 
    \STATE $D = D + \eta \cdot \nabla D(L_{\text{adv}})$  // Maximize $L_{\text{adv}}$
\ENDFOR
\RETURN Trained master key model $G_a$
\end{algorithmic}
\end{algorithm}

The total loss function for training $G_a$ denoted by $L$ is shown in Eq~\eqref{eq4}.

\begin{equation}
\mathcal{L} = \alpha \cdot \mathcal{L}_{adv} + \beta \cdot \mathcal{L}_g + \gamma \cdot \mathcal{L}_{h}
\label{eq4}
\end{equation}

Where $\alpha$, $\beta$  and $\gamma$ are the loss coefficients. By optimizing losses together, $G_a$ model is trained to work as a surrogate $GAN$ model of $AVIH$ secret key model $G$. In other words,  $G_a$ model learns common features of different secret key $\widetilde{G}$ encrypted images $\hat{x}^\prime$, which are preserved in $\hat{x}^\prime$ to reconstruct the $\hat{x}$ using the secret key $\widetilde{G}$. $G_a$ model utilises that to learn the mapping from  $x^\prime$ to  $\hat{x}$ without needing access to the secret key models $G$ used in the $AVIH$ method. Alogirithm~\ref{algo:agan_model} shows how the $G_a$ model is trained using the surrogate dataset $S$ such that the generator can be used to extract the $x$ from the visually encrypted $x^\prime$ in AVIH.

\section{Experiments and results}
\label{sec:blind}
In this section, we validate the effectiveness of our $AGAN$ attack on deep learning-based $PE$ method $AVIH$. Experiments are conducted on a machine with two $14$-$core$ $Intel$ $i9$ $10940X$ $CPU$, $128$ $GB$ $RAM$, and two $Nvidia$ $RTX$-$5000$ $GPU$s with $16$ $GB$ $VRAM$ each.

\subsection{Experimental Setups}
\label{experiment}
\subsubsection{Generating Surrogate Dataset: } We used the $\textit{CelebA}$~\cite{celeba_d1} and $\textit{COCO}$~\cite{coco_d2} datasets to train data $GAN$s $\hat{G_1}$ to $\hat{G_4}$. The $\textit{CelebA}$ dataset includes $203K$ images of the faces of various celebrities. To improve the performance of our proposed attack, we included the $\textit{COCO}$ dataset, which has $328K$ images across different categories like dog, cat, person, $etc$. Images from both datasets were scaled to $112\times112$ to standardise input size for training all the data $GAN$s. Here, we have used four instances of the data $GAN$s (two of them are trained on the $CelebA$ dataset, and the rest of the two are trained on the $COCO$ dataset).  We sampled $50K$ data points from each of the datasets. Among them, $40K$ data samples were set aside for training, and the remaining $10K$ samples were kept for model validation.  The architecture of the data $GAN$s is based on the architecture of $pix2pix$  model~\cite{pix2pix}  with the specific configurations: the generator is based on the $ResNet\_9blocks$ architecture, while the discriminator had three convolutional layers. Additionally, a standard learning rate of $0.0002$ is used across all data $GAN$ models. The settings remained consistent, while the seeds were different for all the data $GAN$ models. We used the $AVIH$ method~\cite{pe6} for training the data $GAN$s to generate the encrypted image from the corresponding original image, as shown in Figure \ref{fig:2}. Once all the data $GAN$s are trained, they are used to generate the surrogate dataset $S$ to train the master key model $G_a$ model. A total of $40K$ surrogate data is generated from four data $GAN$s ($10K$ from each data $GAN$). $30K$ data are used for training $G_a$ while the remaining $10K$ is used for validating $G_a$.

\subsubsection{Evalution Metrics.} To evaluate the effectiveness of our method, we explore the three distinct metrics to quantify the similarity between original and reconstructed images: Cosine Similarity, Structural Similarity Index ($SSIM$), and Learned Perceptual Image Patch Similarity ($LPIPS$). Cosine Similarity measures the directional alignment of pixel vectors of images in a multidimensional space, reflecting overall similarity irrespective of individual pixel intensities. $SSIM$ measures local image patches, incorporating luminance, contrast, and structure to evaluate content structure similarity~\cite{SSIM}. $LPIPS$ utilizes deep learning model to extract texture, color, and shape features, providing a nuanced evaluation of image perceptual similarity~\cite{LPIPS}.

\begin{figure}[!htb]
  \centering
  \includegraphics[scale=0.31]{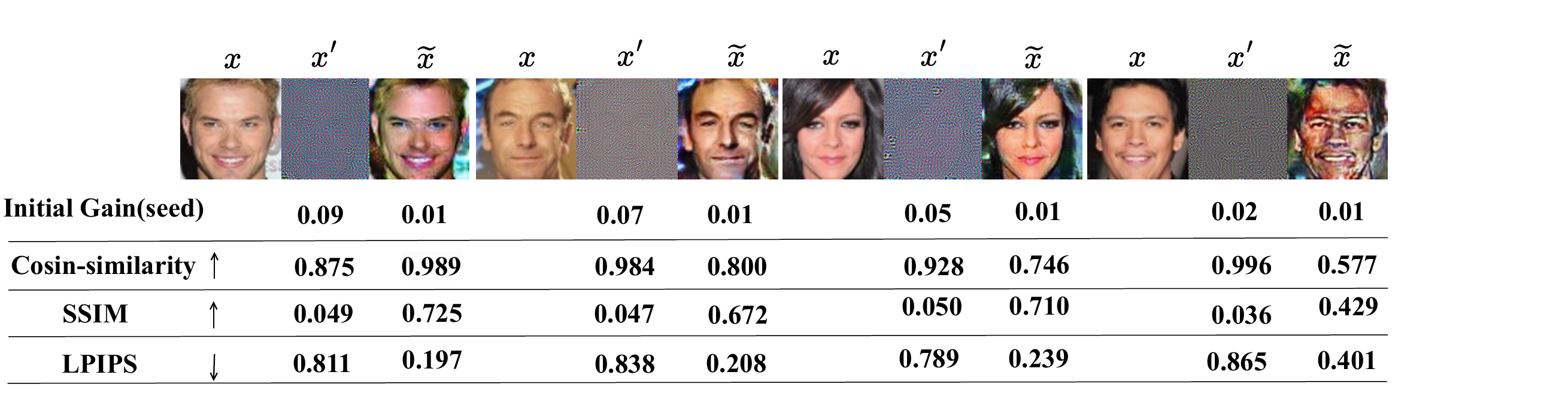}
  \caption{Master key model cosine similarity, SSIM and LPIPS score for reconstructed image $\widetilde{x}$ from encrypted image $x^\prime$ corresponding to the original image $x$. 
  \label{fig:}
  }
\end{figure}

\subsection{$AGAN$ based attack}
\label{att}
In the $AGAN$ method,  we adapt generator and discriminator architectures from the pix2pix~\cite{pix2pix} model. Here, the master key model generator uses $ResNet50$~\cite{resnet} as an encoder, capturing high-level input image representations as a latent feature map. This latent feature map generated by the encoder is then passed to the decoder, which uses it to generate the output image in the target domain. The decoder consists of convolutional layers and upsampling operations to reconstruct the output image at the desired resolution $112\times112$. In the $G_a$ model, we also utilise the pre-trained $ResNet50$  model as the feature extractor. $ResNet50$ is a powerful feature extractor known for capturing universal features that are robust to changes in viewpoint, lighting, and other image aspects.

We trained our $G_a$ model with batches of $32$ images, using a learning rate $0.0002$ for both the generator and discriminator. We trained for $220$ epochs to achieve the optimum performance. The training dataset contains $40K$ images, while $10K$ held out for validation. Once trained, the $G_a$ model is fed with the  $x^\prime$ generated from $AVIH$ and the output/reconstructed image $\widetilde{x}$ from the $G_a$ model is obtained. For this, we collected encrypted data from different instances of $AVIH$, each having a secret $GAN$ initialized with a unique seed. The master key model $G_a$ successfully reconstructed the encrypted images from all the instances of $AVIH$. 

\begin{figure}
  \begin{minipage}[b]{0.48\textwidth}
    \centering
    \includegraphics[scale=0.33]{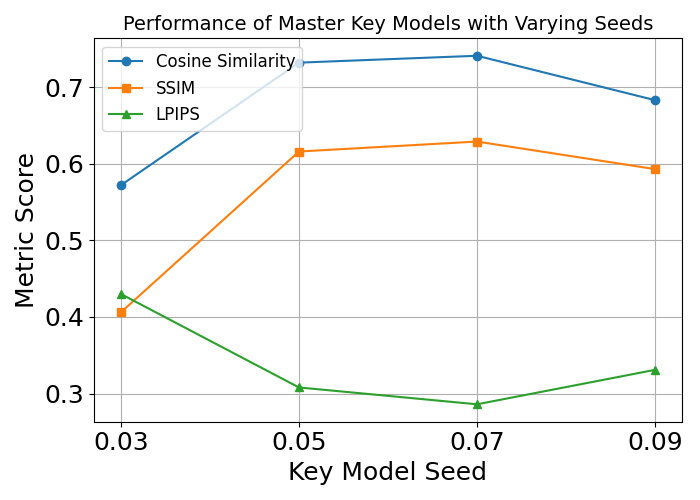}
    \captionof{figure}{Performance of Master key model on different key encrypted images}
    \label{fig:3}
  \end{minipage}\hfill
  \begin{minipage}[b]{0.48\textwidth}
    \centering
    
      \begin{tabular}{|c|c|c|c|}
\hline
\begin{tabular}[c]{@{}c@{}}Initial seed \\ of Secret \\$GAN$ models\end{tabular} & \begin{tabular}[c]{@{}c@{}}cosine-\\ similarity\end{tabular} &  SSIM  & LPIPS \\ \hline
0.02                & 0.572                                & 0.406 & 0.430 \\ \hline
0.05                & 0.732                                & 0.616 & 0.308 \\ \hline
0.07                & 0.741                                & 0.629 & 0.286 \\ \hline
0.09                & 0.683                                & 0.593 & 0.331 \\ \hline
\end{tabular}
    
    \captionof{table}{Quality score of $\widetilde{x}$ corresponding $x^\prime$  to evaluate the overall performance of master key model on 1000 sample data}
    \label{tabel:1}
  \end{minipage}
\end{figure}

The quality of the reconstruction process or breaking the privacy of $AVIH$ by $AGAN$ method is shown in Figure~\ref{fig:},~\ref{fig:3}, and Table~\ref{tabel:1}. Figure~\ref{fig:} shows original image $x$, its $AVIH$ encrypted image $x^\prime$, and the reconstructed image $\widetilde{x}$ by our $AGAN$ method from $x^\prime$. The initial seed value under the images shows the seed of the generator from which the image is encrypted ($G$ for $x^\prime$) or reconstructed (master key model $G_a$ for $\widetilde{x}$). Cosine-similarity, $SSIM$, and $LPIPS$ values under the image show the corresponding relation between the image ($x^\prime$ or $\widetilde{x}$ and  $x$).

Table~\ref{tabel:1} shows the effectiveness of our proposed method on successfully reconstructing $\widetilde{x}$ from $x^\prime$, encrypted by all four instances of $AVIH$ on $1000$ samples. The seed values are given in the first column of Table~\ref{tabel:1}. The reconstructed images have high cosine similarity with their original counterparts, indicating $AGAN$'s ability to preserve essential features successfully. Furthermore, high $SSIM$ values between the original and reconstructed images indicate a substantial structural resemblance between the original and reconstructed images. Along the same line, a low $LPIPS$ score indicates high perceptual similarity between the reconstructed and original images. The performance of $AGAN$ with respect to different seeds and the metric scores are shown in Figure~\ref{fig:3}.

The $AGAN$ method also struggles to reconstruct the original image $x$ from the encrypted image $x^\prime$ in exceptional circumstances where the encrypted image $x^\prime$ of $AVIH$ is generated with precision and the hyperparameters are chosen carefully. In this case, the reconstructed images $\widetilde{x}$ may have a slightly lower cosine similarity than others. Nonetheless, even under such settings, reconstructed image visual quality remained above $60$\%, indicating the master key model's robustness and effectiveness as shown in the $3^{rd}$ column of Figure~\ref{fig:}.

\subsection{Comparison and Effectiveness}
\begin{figure}
  \begin{minipage}[b]{0.49\textwidth}
    \centering
    \includegraphics[scale=0.45]{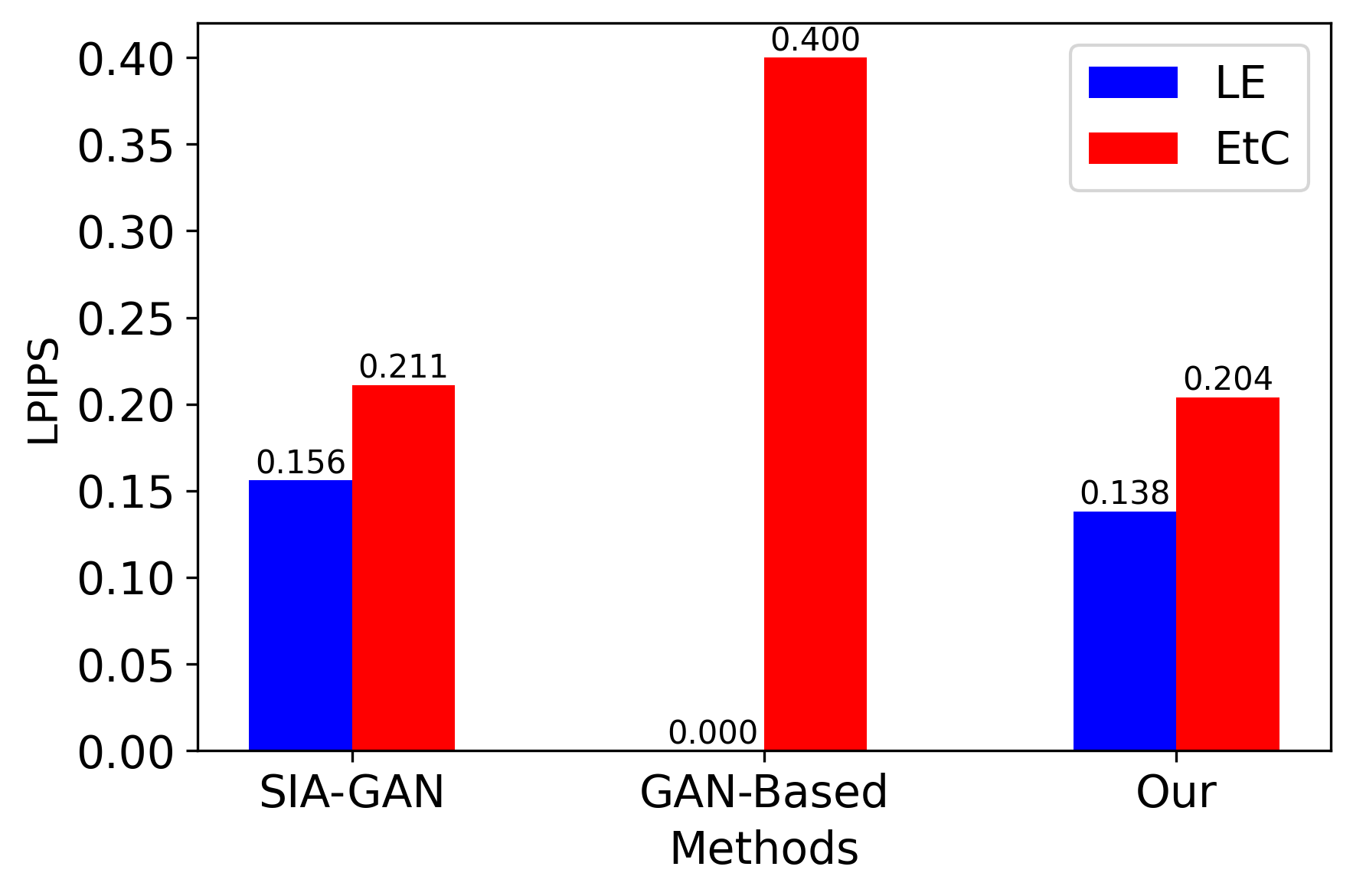}
    \captionof{figure}{LPIPS score based comparison between $SIA-GAN$, $GAN$-Based attack and our attack method to decrypted the tradition $PE$ method $LE$, $EtC$ }
    \label{comp}
  \end{minipage}\hfill
  \begin{minipage}[b]{0.38\textwidth}
    \centering

\begin{tabular}{|c|c|c|}
\hline
Methods     & \begin{tabular}[c]{@{}c@{}}$LE$~\cite{tanaka} \hspace{1cm}\end{tabular} & \begin{tabular}[c]{@{}c@{}}$EtC$~\cite{pe8}\end{tabular} \\ \hline
\begin{tabular}[c]{@{}c@{}}$SIA-GAN$  \\~\cite{GAN_A2}\end{tabular}  & 0.156  & 0.211 \\ \hline
\begin{tabular}[c]{@{}c@{}}$GAN$-Based  \\~\cite{GAN_A1}\end{tabular}  & -     & 0.4   \\ \hline
Our         & 0.138 & 0.204 \\ \hline
\end{tabular}
   \vspace{0.8cm}
    \captionof{table}{Comparision between $SIA-GAN$, $GAN$-Based attack and our attack method based on LPIPS score}
    \label{tabel:2}
  \end{minipage}
\end{figure}

In this section, we perform the $ AGAN$ method on traditional $PE$ methods.  We targeted the existing traditional $PE$ methods, which includes the $LE$ method~\cite{tanaka} and   $EtC$~\cite{pe8,pe4}. In $LE$, a secret key performs block-wise shuffling of color components. In contrast, $Etc$ uses a secret encryption key for each image to perform image transformation and shuffling of color components.  To perform the $AGAN$ method on $LE$, we first created the surrogate dataset using four different keys to encrypt the $CIFAR10$ dataset with a dimension of $32 \times 32$. Then, we collected  $12500$ data sample from each key.   $40K$ data samples are used to train the master key model $G_a$, and  $10K$ data points are used for validation. To train the master key model $G_a$ on the $EtC$ method, we have taken $50K$ data points from the $CIFAR10$ dataset with a dimension of $32 \times 32$ with different keys to encrypt each image. $40K$ data points are used for training the master key model $G_a$ of $AGAN$ method, and $10K$ data points are used for validate $G_a$. We have used the same architecture and hyperparameters of the master key model $G_a$ as mentioned in Section~\ref{att}. Once trained, a different key encrypted image $x^\prime$ is fed to the master key $G_a$ model, which reconstructs the original image $x$ for the given encrypted image $x^\prime$. To show the robustness of our $G_a$ against traditional $PE$, we demonstrated the $LPIPS$ values between the original image $x$,  and the reconstructed image $\widetilde{x}$  corresponding given encrypted image $x^\prime$ are shown in Table~\ref{tabel:2}.

To show the efficacy of our $AGAN$ method, we compared its performance with the exiting attack method of $SIA-GAN$~\cite{GAN_A2} and the $GAN$-based attack~\cite{GAN_A1} for $PE$ method of the $LE$ and $EtC$. Our attack successfully reconstructed the original content from encrypted images $x^\prime$ in both encryption methods. We take $1000$ data from each encryption method to validate the model, which is unknown to our trained model. The result is shown in Table~\ref{tabel:2} and Figure~\ref{comp}, where we can see that our attack has a small $LPIPS$ score compared to existing attack methods, which indicates that our attack outperforms the existing attack methods.

\begin{figure}[!htb]
  \centering
  \includegraphics[scale=0.45]{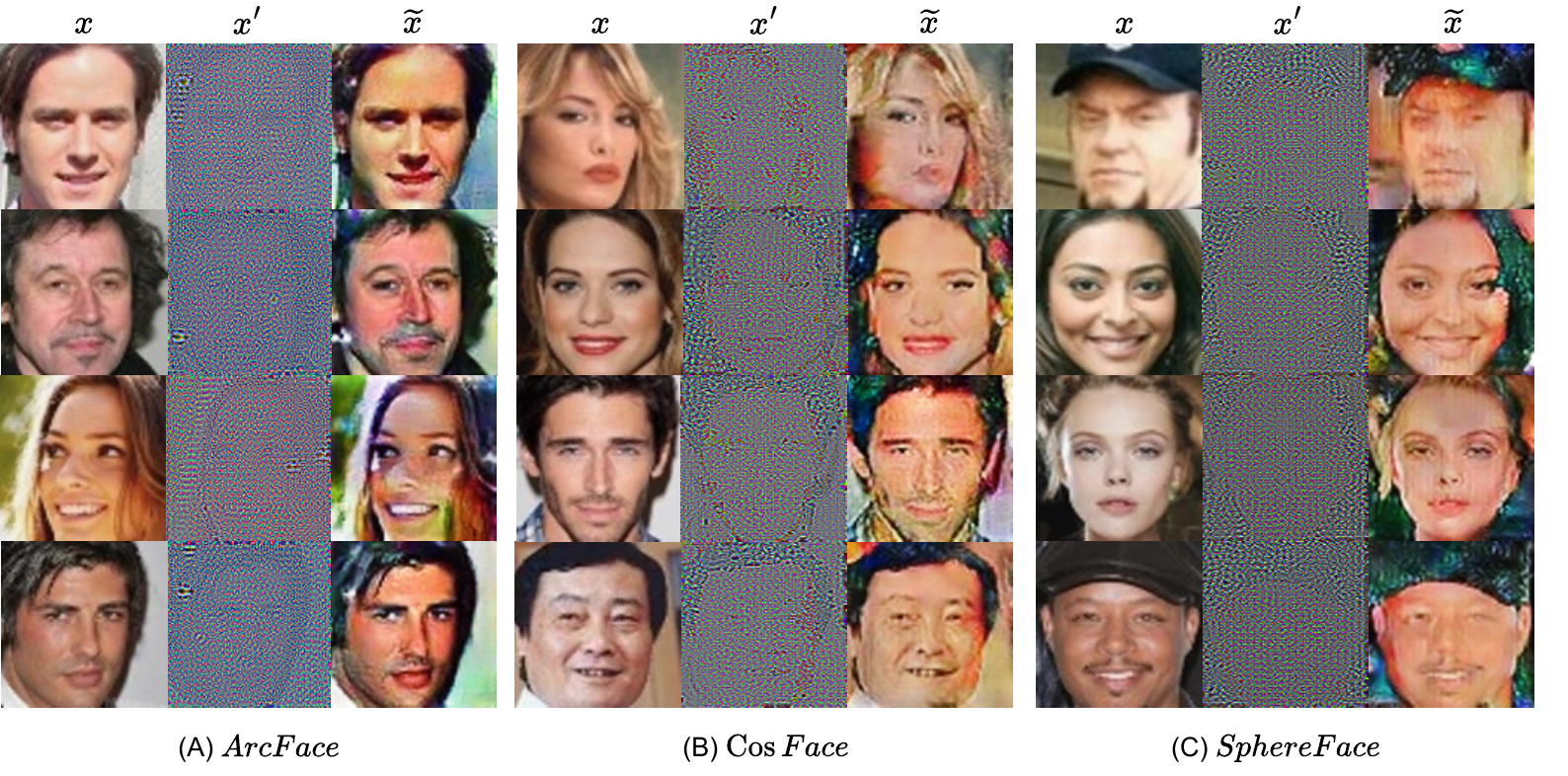}
  \caption{Master key model $G_a$ trained over the $ArcFace$ using surrogate dataset $S$ generalization over the $CosFace$ and the $SphereFace$ model . 
  \label{fig:4}
  }
\end{figure}

We also explored the transferability of the $AGAN$ method by showing that if an adversary develops attacks for a particular service model, it will also work for other models with a similar task. This shows the transferability of our $AGAN$ method. The result is shown in Figure~\ref{fig:4}, where we attack the $ArcFace$ model by training the master key model $G_a$ of $AGAN$, which reconstructs any secret $GAN$ encrypted image $x^\prime$ for the $ArcFace$ model and also works with the $CosFace$ and the $SphereFace$ model.

\section{Conclusion}
In this paper, we performed a detailed security analysis of $SOTA$ deep learning-based $PE$ method $AVIH$. We developed a new security evaluation tool referred to as the $AGAN$ attack. $AGAN$ shows $SOTA$ deep learning-based $PE$ method $AVIH$ is not secure. The proposed $AGAN$ creates the master key model, which is flexible and easily transferable to alternative service models with similar tasks. The adaptability of $AGAN$ across diverse $PE$ methods is also shown by successfully attacking the traditional methods of $LE$  and  $EtC$.  $AGAN$ can be used as an effective security evaluation tool by the designer of the $PE$ method. Its effective use will help in designing secure $PE$ methods in future.


\clearpage  

%
%
\bibliographystyle{splncs04}
\bibliography{main}

\begin{thebibliography}{10}
\providecommand{\url}[1]{\texttt{#1}}
\providecommand{\urlprefix}{URL }
\providecommand{\doi}[1]{https://doi.org/#1}

\bibitem{he5}
Aono, Y., Hayashi, T., Phong, L.T., Wang, L.: Privacy-preserving logistic regression with distributed data sources via homomorphic encryption. IEICE TRANSACTIONS on Information and Systems  \textbf{99}(8),  2079--2089 (2016)

\bibitem{he6}
Aono, Y., Hayashi, T., Wang, L., Moriai, S., et~al.: Privacy-preserving deep learning via additively homomorphic encryption. IEEE transactions on information forensics and security  \textbf{13}(5),  1333--1345 (2017)

\bibitem{arockiam2014efficient}
Arockiam, L., Monikandan, S.: Efficient cloud storage confidentiality to ensure data security. In: 2014 International Conference on Computer Communication and Informatics. pp.~1--5. IEEE (2014)

\bibitem{baevski2020wav2vec}
Baevski, A., Zhou, Y., Mohamed, A., Auli, M.: wav2vec 2.0: A framework for self-supervised learning of speech representations. Advances in neural information processing systems  \textbf{33},  12449--12460 (2020)

\bibitem{cipat}
Chang, A.H., Case, B.M.: Attacks on image encryption schemes for privacy-preserving deep neural networks. arXiv preprint arXiv:2004.13263  (2020)

\bibitem{chen2012data}
Chen, D., Zhao, H.: Data security and privacy protection issues in cloud computing. In: 2012 international conference on computer science and electronics engineering. vol.~1, pp. 647--651. IEEE (2012)

\bibitem{pe4}
Chuman, T., Sirichotedumrong, W., Kiya, H.: Encryption-then-compression systems using grayscale-based image encryption for jpeg images. IEEE Transactions on Information Forensics and security  \textbf{14}(6),  1515--1525 (2018)

\bibitem{pe3}
Ding, Y., Wu, G., Chen, D., Zhang, N., Gong, L., Cao, M., Qin, Z.: Deepedn: A deep-learning-based image encryption and decryption network for internet of medical things. IEEE Internet of Things Journal  \textbf{8}(3),  1504--1518 (2020)

\bibitem{resnet}
He, K., Zhang, X., Ren, S., Sun, J.: Deep residual learning for image recognition. corr abs/1512.03385 (2015) (2015)

\bibitem{he2016deep}
He, K., Zhang, X., Ren, S., Sun, J.: Deep residual learning for image recognition. In: Proceedings of the IEEE conference on computer vision and pattern recognition. pp. 770--778 (2016)

\bibitem{pix2pix}
Isola, P., Zhu, J.Y., Zhou, T., Efros, A.A.: Image-to-image translation with conditional adversarial networks. arxiv e-prints. arXiv preprint arXiv:1611.07004  (2016)

\bibitem{pe1}
Ito, H., Kinoshita, Y., Aprilpyone, M., Kiya, H.: Image to perturbation: An image transformation network for generating visually protected images for privacy-preserving deep neural networks. IEEE Access  \textbf{9},  64629--64638 (2021)

\bibitem{inverse}
Ito, H., Kinoshita, Y., Kiya, H.: Image transformation network for privacy-preserving deep neural networks and its security evaluation. In: 2020 IEEE 9th Global Conference on Consumer Electronics (GCCE). pp. 822--825. IEEE (2020)

\bibitem{pe2}
Kiya, H., Maung, A.P.M., Kinoshita, Y., Imaizumi, S., Shiota, S., et~al.: An overview of compressible and learnable image transformation with secret key and its applications. APSIPA Transactions on Signal and Information Processing  \textbf{11}(1) (2022)

\bibitem{coco_d2}
Lin, T.Y., Maire, M., Belongie, S., Hays, J., Perona, P., Ramanan, D., Doll{\'a}r, P., Zitnick, C.L.: Microsoft coco: Common objects in context. In: Computer Vision--ECCV 2014: 13th European Conference, Zurich, Switzerland, September 6-12, 2014, Proceedings, Part V 13. pp. 740--755. Springer (2014)

\bibitem{celeba_d1}
Liu, Z., Luo, P., Wang, X., Tang, X.: Deep learning face attributes in the wild. In: Proceedings of the IEEE international conference on computer vision. pp. 3730--3738 (2015)

\bibitem{he4}
Lou, Q., Jiang, L.: She: A fast and accurate deep neural network for encrypted data. Advances in neural information processing systems  \textbf{32} (2019)

\bibitem{GAN_A2}
Madono, K., Tanaka, M., Onishi, M., Ogawa, T.: Sia-gan: Scrambling inversion attack using generative adversarial network. IEEE Access  \textbf{9},  129385--129393 (2021)

\bibitem{GAN_A1}
MaungMaung, A., Kiya, H.: Generative model-based attack on learnable image encryption for privacy-preserving deep learning. arXiv preprint arXiv:2303.05036  (2023)

\bibitem{simonyan2014very}
Simonyan, K., Zisserman, A.: Very deep convolutional networks for large-scale image recognition. arXiv preprint arXiv:1409.1556  (2014)

\bibitem{singh2017cloud}
Singh, A., Chatterjee, K.: Cloud security issues and challenges: A survey. Journal of Network and Computer Applications  \textbf{79},  88--115 (2017)

\bibitem{pe5}
Sirichotedumrong, W., Kinoshita, Y., Kiya, H.: Pixel-based image encryption without key management for privacy-preserving deep neural networks. Ieee Access  \textbf{7},  177844--177855 (2019)

\bibitem{pe7}
Sirichotedumrong, W., Kiya, H.: A gan-based image transformation scheme for privacy-preserving deep neural networks. In: 2020 28th European Signal Processing Conference (EUSIPCO). pp. 745--749. IEEE (2021)

\bibitem{pe8}
Sirichotedumrong, W., Maekawa, T., Kinoshita, Y., Kiya, H.: Privacy-preserving deep neural networks with pixel-based image encryption considering data augmentation in the encrypted domain. In: 2019 IEEE International Conference on Image Processing (ICIP). pp. 674--678. IEEE (2019)

\bibitem{pe6}
Su, Z., Zhou, D., Wang, N., Liu, D., Wang, Z., Gao, X.: Hiding visual information via obfuscating adversarial perturbations. In: Proceedings of the IEEE/CVF International Conference on Computer Vision. pp. 4356--4366 (2023)

\bibitem{tanaka}
Tanaka, M.: Learnable image encryption. In: 2018 IEEE International Conference on Consumer Electronics-Taiwan (ICCE-TW). pp.~1--2 (2018). \doi{10.1109/ICCE-China.2018.8448772}

\bibitem{vaswani2017attention}
Vaswani, A., Shazeer, N., Parmar, N., Uszkoreit, J., Jones, L., Gomez, A.N., Kaiser, {\L}., Polosukhin, I.: Attention is all you need. Advances in neural information processing systems  \textbf{30} (2017)

\bibitem{he7}
Wang, Y., Lin, J., Wang, Z.: An efficient convolution core architecture for privacy-preserving deep learning. In: 2018 IEEE International Symposium on Circuits and Systems (ISCAS). pp.~1--5. IEEE (2018)

\bibitem{SSIM}
Wang, Z., Bovik, A.C., Sheikh, H.R., Simoncelli, E.P.: Image quality assessment: from error visibility to structural similarity. IEEE transactions on image processing  \textbf{13}(4),  600--612 (2004)

\bibitem{LPIPS}
Zhang, R., Isola, P., Efros, A.A., Shechtman, E., Wang, O.: The unreasonable effectiveness of deep features as a perceptual metric. In: Proceedings of the IEEE conference on computer vision and pattern recognition. pp. 586--595 (2018)

\end{thebibliography}
\end{document}